%% file: main.tex
\documentclass[10pt,twocolumn,letterpaper]{article}
\usepackage[T1]{fontenc}
\usepackage[pagenumbers]{cvpr}      
\input{preamble}

\usepackage{booktabs}
\usepackage{multirow}
\usepackage{amsmath,amssymb}
\usepackage{subcaption}

\definecolor{cvprblue}{rgb}{0.21,0.49,0.74}
\usepackage[pagebackref,breaklinks,colorlinks,allcolors=cvprblue]{hyperref}

\def\confName{CVPR}
\def\confYear{2027}

\usepackage{xspace}
\newcommand{\method}{Z2H\xspace}

\newcommand{\chk}{{\color{green}$\checkmark$}}
\newcommand{\xmrk}{{\color{red}$\times$}}
\newcommand{\hathatYR}{$\widehat{\widehat{Y}}$\raisebox{1.5mm}{$\scriptstyle R$}}
\newcommand{\sectitle}[1]{\noindent{\bf #1}}

\title{Zero to Hero: 6DoF Pose Estimation Through \\ Cross-Domain Hard Ensemble Distillation}
\author{Naser Al-Obeidat \qquad Ali Rafiaei \qquad Michael Greenspan\\
{\small\it Dept. Electrical and Computer Engineering, Ingenuity Labs, Queen's University, Kingston, ON, Canada}\\
{\small\tt \{21nao3, 23rppc, michael.greenspan\}@queensu.ca}\\[-1.2em]
}
\begin{document}
\raggedbottom

\maketitle

\input{sec/0_abstract}

\input{figs/bannerfigure.tex}
\input{sec/1_intro}

\input{sec/2_related}

\input{sec/3_method}
\input{sec/4_experiments}
\input{sec/5_conclusion}

\clearpage
{\small
\bibliographystyle{ieeenat_fullname}
\bibliography{main}
}

\end{document}

%% file: preamble.tex
\usepackage{currfile} 
\usepackage{caption}

%% file: sec/0_abstract.tex
\begin{abstract}
Supervised 6DoF object pose estimators provide fast inference but require pose annotations or face a synthetic-to-real domain gap. Zero-shot models can estimate poses for unseen objects without object-specific training, but their inference cost limits deployment. We introduce Zero to Hero (Z2H), a cross-domain hard ensemble distillation framework that uses zero-shot and simulation-trained teachers to label real RGB-D images without real pose annotations. Candidate poses are refined against observed depth, filtered by a pretrained binary verifier, and selected by minimum Chamfer distance. An RGB student is then trained on the resulting silver pseudo-labels together with labeled synthetic images, without modifying its architecture or inference procedure. Across the seven core BOP datasets, ZebraPose improves from 72.0 to 75.4 AR without test-scene images and to 77.7 AR when adaptation also uses unlabeled non-target frames from test scenes. In the latter setting, SurfEmb and GDRNPP reach 70.6 and 74.3 AR, exceeding their listed real-ground-truth-supervised counterparts by 3.3 and 1.5 AR points. Under the Self6D++ protocol on LM-O, Z2H reaches 70.4 ADD(-S). Ablations show that refinement and verification make pseudo-labels more effective, while a matched-image comparison finds that the advantage over ground-truth supervision depends on the dataset. Z2H retains RGB-only student inference; depth and teachers are used only to generate training labels. We  release both the BOP silver dataset and the Z2H source code to the public domain upon publication of this work.
\end{abstract}

%% file: figs/bannerfigure.tex
\begin{figure*}[t]
\centering
\includegraphics[width=1.0\textwidth]{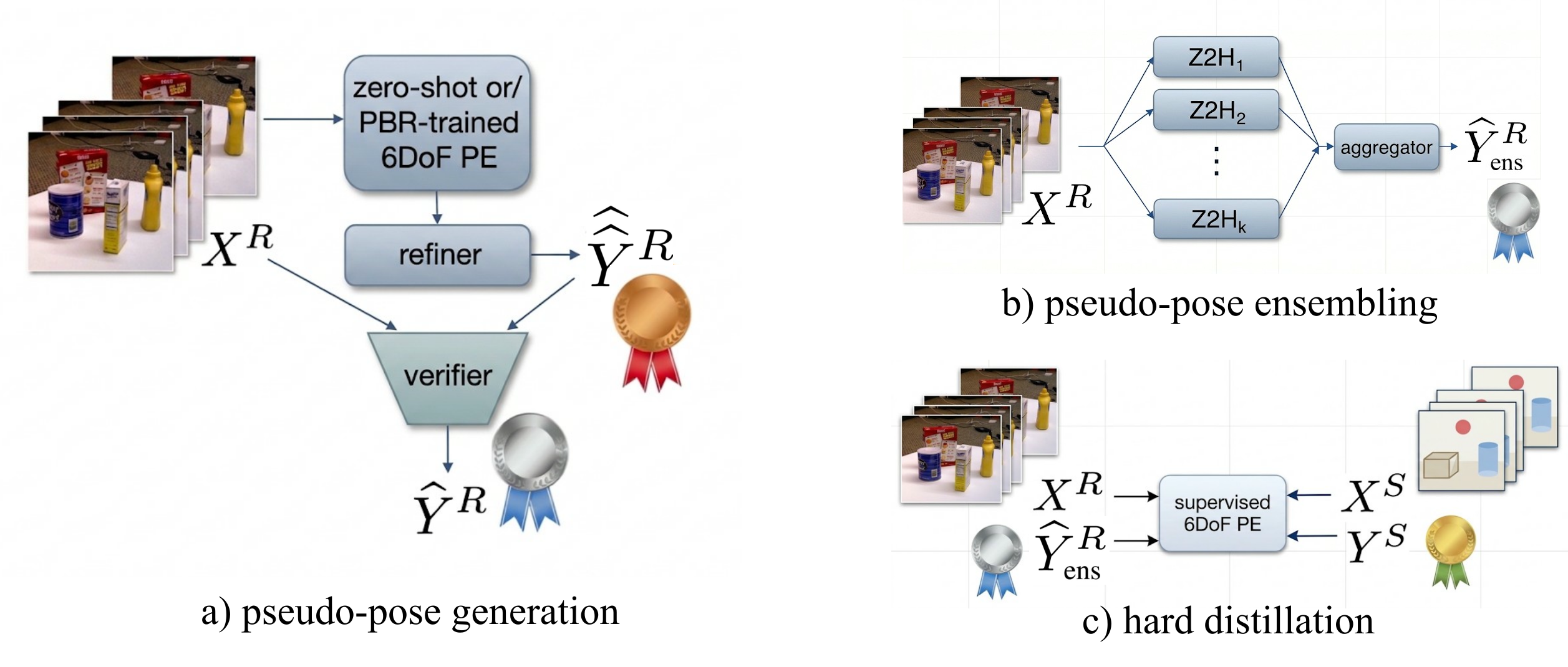}
\caption{
\textbf{Overview of Z2H.}
a) Pseudo-pose generation.
$X^R = $ unlabeled real images,
\hathatYR$=$
noisy (bronze) pseudo-poses,
$\widehat{Y}^R\!\!=$
cleaned (silver) pseudo-poses.
b) Pseudo-pose ensembling. 
$\widehat{Y}^R_{\mathrm{ens}}$ = aggregation of pseudo-poses
output from multiple Z2H modules.
c) Cross-domain hard distillation. 
Student model is supervised with
real images $X^R$ and
corresponding
ensemble (silver) pseudo-poses
$\widehat{Y}^R_{\mathrm{ens}}$,
combined with 
synthetic images $X^S$
with ground truth (gold) poses 
$Y^S$.
}
\label{fig:banner}
\end{figure*}

%% file: sec/1_intro.tex
\section{Introduction}
\label{sec:intro}
Learning-based methods substantially improved 6DoF pose estimation (6DoF PE) \cite{xiang2017posecnn, wang2019densefusion, kehl2017ssd6d, peng2019pvnet, li2018deepim, park2019pix2pose, rcvpose2022}. This improvement came at the cost of requiring supervision, involving both expensive dataset
labeling and at times lengthy training phases. Automatically labeled simulated datasets relieved the cost of manual labeling~\cite{tremblay2018falling}, although a domain gap persists when compared against training on real labeled data \cite{hodan2020bop}. A variety of self- and semi-supervised approaches have been developed that  reduce, but not completely eliminate, this domain gap \cite{wang2020self6d, wu2024rkhspose},
and real pose annotations remain an important source of supervision.

Recent zero-shot approaches have been proposed to remove the need for annotated data for unseen objects, essentially revisiting the  formulation of the problem that was once standard \cite{nguyen2023cnos}. These methods mirror the previous approach of establishing correspondences between local features, which are now much more discriminating as they are learned from pretrained vision transformers \cite{dosovitskiy2020image, oquab2025dinov3}. Zero-shot methods provide a different accuracy and runtime trade-off from instance-specific supervised models \cite{rafiaei2026b2tfpose, wen2024foundationpose, labbe2022megapose}, and while they avoid object-specific pose annotation and training, inference is relatively slow, requiring on the order of seconds per object compared with a few tens of milliseconds for a supervised model.
The above considerations lead to the following research question: \\

\textit{Is there an approach to 6DoF PE that combines the annotation-free generality of zero-shot methods, with the high accuracy and fast inference of supervised methods?} \\

Our solution, Zero to Hero (Z2H), is to use a 
combination of zero-shot 
and
simulation-trained
methods 
to automatically annotate an unlabeled real dataset with pseudo-poses, which are then aggregated to train a supervised method. This process is a form of transfer learning known as \textit{cross-domain ensemble distillation} \cite{hinton2015distilling, radosavovic2018data}, and is illustrated in Fig.~\ref{fig:banner}. 
We explicitly store the pseudo-poses output from the ensemble module, 
which produces reusable pseudo-labels for the adapted real datasets
and is a form of 
\textit{hard distillation}.
The performance of Z2H relies on the introduction of a verification module, which is a binary classifier that cleans the pseudo-poses output from the zero-shot method by detecting and discarding false positives. Our experimental results show that Z2H boosts performance of models supervised with simulated data  
with the results reported separately for adaptation with and without test-scene images.
Our contributions are:
\begin{itemize}
    \item Z2H, a novel approach to hard distillation of 6DoF PE methods that requires no real data annotation. The Z2H method includes a pseudo-pose verification module that improves the reliability of its pseudo-poses,
    and a pose ensemble module that matches unioning and outperforms averaging. 
    \item Extensive experiments on the seven core BOP datasets \cite{hodan2018bop} that demonstrate the performance benefits of Z2H,
    which improves three RGB students without real pose annotations. Comparisons with supervised methods distinguish adaptation with and without access to unlabeled test-scene frames. 
    These experiments combine two zero-shot models and three supervised models, with different architectures. Ablations and auxiliary experiments identify when refinement, verification and ensembling help.
    \item Pseudo-labels generated by Z2H for the five adapted core BOP datasets, which we call \textit{BOP silver}. We  release both the BOP silver dataset and the Z2H source code to the public domain upon publication of this work.
\end{itemize}

%% file: sec/2_related.tex
\section{Related Work}
\label{sec:related}

\sectitle{Instance-level 6DoF PE.}
Correspondence-based methods regress 2D--3D correspondences and solve PnP with RANSAC~\cite{peng2019pvnet,park2019pix2pose,li2019cdpn,su2022zebrapose,haugaard2022surfemb,hodan2020epos}, whereas direct methods regress the pose end-to-end~\cite{xiang2017posecnn,wang2021gdrnet,liu2025gdrnpp}, and refinement methods iterate a render-and-compare process~\cite{li2018deepim,labbe2020cosypose,hu2022pfa}.
GDRNPP~\cite{liu2025gdrnpp} combines geometry-guided direct regression with a strong detector and provides a baseline among RGB methods trained on synthetic data.
\method is independent of which type of instance-level 6DoF PE method is used; we demonstrate it on a direct-regression network (GDRNPP), a dense-embedding network (SurfEmb), and a surface-encoding network (ZebraPose).

\sectitle{Synthetic-to-real adaptation without annotations.}
Self6D~\cite{wang2020self6d} and Self6D++~\cite{wang2021self6dpp} self-supervise a 6DoF PE network on unlabeled real RGBD images through differentiable rendering and a noisy-student scheme.
DSC-PoseNet~\cite{yang2021dscposenet} adapts with weak 2D supervision, while TexPose~\cite{chen2023texpose} learns real textures for re-rendering, PseudoFlow~\cite{hai2023pseudoflow} refines pseudo-labels with optical flow, MSDA~\cite{lin2024msda} adapts monocular models across domains, and ONDA-Pose~\cite{tan2025ondapose} adapts with occlusion-aware neural rendering of pseudo-labeled real images.
Alternately, sim-to-real self-training for bin picking~\cite{chen2022sim2realbin} iteratively selects reliable student predictions as pseudo-labels, and SLAM-supported self-training~\cite{lu2022slamselftrain} makes pseudo-labels consistent across frames through pose-graph optimization. For example, 
SMOC-Net~\cite{tan2023smocnet} and RKHSPose~\cite{wu2024rkhspose} learn from relative camera poses and pseudo-keypoints, respectively.
These methods draw supervision from the student itself or from rendering losses.

In contrast, \method adds an independent zero-shot teacher, whose predictions can complement the student's, and lets the observed geometry arbitrate between the two.
Following Self6D++, we use unlabeled real depth only at training time and evaluate as an RGB method.

\sectitle{Zero-shot and novel-object PE.}
CNOS~\cite{nguyen2023cnos} segments novel objects by matching rendered templates; MegaPose~\cite{labbe2022megapose}, FoundationPose~\cite{wen2024foundationpose}, GenFlow~\cite{moon2024genflow}, SAM-6D~\cite{lin2024sam6d} and 
recently B2TFPose~\cite{rafiaei2026b2tfpose} estimate unseen object poses directly from CAD models.
These methods trade accuracy and speed for generality.
OSSID~\cite{gu2022ossid} uses a slow zero-shot pose estimator to self-supervise a fast object detector.
\method shares the premise that a slow zero-shot method provides supervision without manual real-pose labels, but then adapts a pose estimator.

%% file: sec/3_method.tex
\section{Method}
\label{sec:method}
Z2H turns unlabeled real RGB-D images into training data for an RGB pose estimator (\cref{fig:banner}). It separates label generation from student deployment: depth, refinement and verification are used offline, while the deployed student retains its original RGB input and architecture.

\sectitle{Pseudo-pose generation.}
Each of $K$ teacher modules receives an unlabeled real image $X^R$ and proposes an object pose. A teacher may be a zero-shot estimator or a supervised estimator pretrained on synthetic images produced by physically based rendering (PBR). We call its raw predictions bronze pseudo-poses, denoted by \hathatYR. Each candidate is refined against the observed depth and passed to a binary verifier. Candidates accepted by the verifier become silver pseudo-poses $\widehat{Y}^R$. This filtering is applied before candidates from different teachers are combined, so a rejected estimate cannot become a training label merely because it is the best among poor alternatives.

\sectitle{Pseudo-pose verification.}
The verifier distinguishes plausible pose estimates from false positives. ValidNet~\cite{mazumder2021multistream} and MegaPose~\cite{labbe2022megapose} use pose verification following estimation; Z2H applies it to curate supervision. We use the pretrained MegaPose verifier, avoiding training a new verifier on real pose labels for the target objects.

The verifier's training formulation compares image pairs $\{x_T,x_{\widehat T}\}$ associated with a ground-truth pose $T$ and a perturbed pose $\widehat T$. Small perturbations are positive examples and larger perturbations are negative examples. A binary cross-entropy objective trains a scalar logit, which is thresholded to accept or reject a candidate. In Z2H, this pretrained classifier filters the refined teacher predictions. Refinement and verification serve different purposes: refinement adjusts an estimate, whereas verification determines whether it should supervise the student. A geometrically refined estimate can still be incorrect, which motivates testing both components separately (\cref{sec:ablations}).

\sectitle{Geometric ensemble selection.}
Different teachers can propose different poses for the same object and image. For every verified candidate, we transform the object CAD model by the estimated pose and compute its Chamfer distance to the observed depth. The candidate with the smallest distance is retained in the ensemble silver dataset, denoted by $\widehat{Y}^R_{\mathrm{ens}}$. Instances without a verified candidate provide no real pseudo-label. This is selection among hard pose estimates, rather than averaging teacher distributions or directly averaging rotations and translations.

Teacher modules may run sequentially or in parallel; ensembling therefore adds offline computation even though it requires no manual real-pose annotation. Minimum-distance selection stores at most one candidate per object and image, whereas taking the union can retain several. We compare these alternatives with pose averaging in \cref{tab:ensemble}.

\sectitle{Cross-domain hard distillation.}
The student is supervised jointly by synthetic RGB images and their ground-truth poses $\{X^S,Y^S\}$ and by real RGB images with selected silver poses $\{X^R,\widehat{Y}^R_{\mathrm{ens}}\}$. Only the RGB channels of $X^R$ are passed to the student. The pose labels are stored explicitly, making the procedure hard distillation and allowing the silver dataset to be reused. Neither teacher inference nor depth processing is required at deployment. We apply the same framework to GDRNPP, SurfEmb and ZebraPose, with no architectural changes to these students.

%% file: sec/4_experiments.tex
\section{Experiments}
\label{sec:experiments}
\input{sec/table_1}

\subsection{Protocol}
\sectitle{Benchmark.}
We follow the RGB protocol of GDRNPP~\cite{liu2025gdrnpp}, i.e. the seven core BOP datasets~\cite{hodan2024bop}, the BOP19 test targets, the official BOP toolkit, and the average recall $\mathrm{AR}\!=\!(\mathrm{AR_{VSD}}\!+\!\mathrm{AR_{MSSD}}\!+\!\mathrm{AR_{MSPD}})/3$.
Published baseline results in \cref{tab:main} follow the cited benchmark sources~\cite{liu2025gdrnpp}.
For our runs, we start from the released PBR checkpoints and PBR-trained detections, with the gains in \cref{tab:main} measured against the published baseline values.
Unless otherwise stated, reported results of \method are averaged over three different fine-tuning seeds.

\sectitle{Unlabeled data.}
We distinguish between two settings. In the first, no images from the test scenes are used for adaptation. For YCB-V and TUD-L, $X^R$ consists of real training images without annotations. For LM-O and T-LESS, adaptation uses real images outside the test scenes. IC-BIN retains the synthetic-only result because no real images outside its test scenes are available.
In the second setting, we
follow the domain adaptation protocol of 
Self6D++~\cite{wang2021self6dpp}, also used by PseudoFlow~\cite{hai2023pseudoflow} and ONDA-Pose~\cite{tan2025ondapose}, 
and allow some unlabeled images from the test partition,
known as \textit{non-target frames},
to be used for pseudo-pose generation.
Specifically, LM-O uses 1,014 non-target frames, and T-LESS and IC-BIN also use non-target frames from their test scenes. ITODD and HB retain each student's synthetic-only leaderboard values in both settings. In either setting, no BOP19 evaluation-target image is used for training, training-label selection or model selection.  Real depth is used only during pseudo-label generation,
so that student supervision and inference are purely RGB-only. Full per-dataset results for both ZebraPose settings are given in Supp. Tab. S3.

\subsection{Comparison with state of the art}

\sectitle{Comparison with supervised baselines.}
\Cref{tab:main} separates synthetic-only methods, methods using real ground-truth labels, and Z2H. The Z2H rows combine FoundationPose and PBR-trained teacher predictions, then train each student on synthetic ground truth and selected real pseudo-labels. GDRNPP, SurfEmb and ZebraPose were chosen because their implementations support retraining.

Without test-scene images ($D$), ZebraPose reaches 75.4 AR, 3.4 points above its synthetic-only EffNet-B4 baseline (72.0) and 0.5 above the listed real-GT ZebraPoseSAT result (74.9). With unlabeled non-target test-scene frames ($T$), it reaches 77.7 AR, a gain of 5.7 over the synthetic-only baseline and 2.8 over ZebraPoseSAT. This score is 0.4 above HccePose(BF)'s 77.3, but the methods do not have matched scene access. We therefore report the two Z2H settings separately rather than treating them as equivalent supervision budgets.

In setting $T$, SurfEmb and GDRNPP reach 70.6 and 74.3 AR. These improve their synthetic-only baselines by 5.9 and 4.1 points, and their listed real-GT counterparts by 3.3 and 1.5 points. All gains are absolute AR-point differences, not relative percentages. The comparison is across published baseline configurations; the controlled experiment in Supplement Table S2 instead holds real images and instances fixed. In that experiment, pseudo-label supervision exceeds GT supervision for all three students on YCB-V but remains below it on TUD-L. Qualitative examples appear in \cref{fig:qual}.

\input{figs/qualitative_results}
\input{sec/table_new_comparisons}

\sectitle{Comparison with label-free domain adaptation.}
\cref{tab:labelfree}
compares Z2H against other domain adaptation methods,
which train on labeled synthetic images 
and unlabeled real images.
Under the Self6D++ protocol on LM-O, \method 
achieves the highest score among the compared methods,
achieving
70.4 ADD(-S), which is
+1.1 above the next highest scoring
ONDA-Pose.
A full comparison is provided in Supp. Tab. S1.

\subsection{Ablation Experiments}
\label{sec:ablations}
\input{sec/ablation_table}

\input{sec/supp_relabel}

A series of ablation experiments were executed to
isolate the impact of
verification and
different refinement modules,
shown in \cref{tab:ablation}.
These experiments used B2TFPose as the zero-shot Z2H teacher,
optionally ensembled with PBR-trained ZebraPose,
and distilled to a ZebraPose student.

A few refinement iterations 
(18 for ICP, 5 for FP)
were executed on the
pseudo-poses output from the Z2H teacher, testing
both ICP and FP refinement
to establish any performance difference between them.
All training also included 
GT labeled PBR images.

\sectitle{Baseline (trial 0).}
Trial 0 is the ZebraPose student trained only on labeled PBR images, with no real images or domain adaptation, and is the baseline for all other trials.

\sectitle{Distillation with raw pseudo-poses (trials 1, 4, 7).}
Here the real-data supervision uses raw pseudo-poses without refinement or verification, alongside the synthetic ground-truth data, 
generated from the zero-shot teacher or the PBR-trained student or their ensemble.
Raw zero-shot and PBR pseudo-poses improve TUD-L by +1.5 and +2.2 AR, but lower LM-O by -1.6 and -1.8 AR and YCB-V by -2.2 and -2.1 AR, respectively.
Only the raw ensemble improves all three datasets, by +1.0, +5.1 and +5.0 AR on LM-O, YCB-V and TUD-L.
Without refinement and verification, the individual teachers' pseudo-poses are too noisy to improve performance on LM-O and YCB-V.

\sectitle{+ Refinement (trials 2, 5, 8).}
Refining every raw pseudo-pose adds +0.4 to +9.3 AR with ICP.
FoundationPose refinement adds more on LM-O and YCB-V, +1.1 to +13.6 AR, with the largest gains for the zero-shot teacher (trial 2: $+6.4$ on LM-O and $+13.6$ on YCB-V).
On TUD-L, FoundationPose refinement improves the individual zero-shot and PBR candidates by +6.4 and +0.4 AR respectively, but lowers the ensemble result by -5.1 AR (trial 8 vs. trial 7), whereas ICP raises AR in every case.

\sectitle{+ Verification (trials 3, 6, 9).}
With ICP refinement, verification improves every configuration on every dataset by 0.2 to 2.7 AR. With FoundationPose refinement, gains range from 0.2 to 8.6 AR except for zero-shot-only YCB-V, which decreases by 0.5. For the ensemble on TUD-L, verification raises the FoundationPose-refined result from 75.7 to 84.3 AR, recovering the refinement loss and exceeding the raw ensemble's 80.8 by 3.5 points. Verification is therefore useful across these configurations, but does not improve every individual result.

\sectitle{+ Ensembling (trials 3, 6, 9, 10).}
With ICP, the verified ensemble (trial 9) exceeds both individual teachers on all three datasets. With FoundationPose refinement, the zero-shot-only teacher leads on LM-O and YCB-V (78.5 and 83.9), while the ensemble leads on TUD-L (84.3). Replacing B2TFPose with FoundationPose as the zero-shot teacher in trial 10 yields 78.8, 84.1 and 86.8 AR, improving trial 9 with the same refiner by 1.0, 0.5 and 2.5 points. These are the highest scores among the refined and verified configurations, although unverified trial 2 is slightly higher on YCB-V (84.4).

\sectitle{FoundationPose versus ICP refinement (trials 3, 6, 9).}
For verified zero-shot, PBR and ensemble teachers, FoundationPose exceeds ICP on LM-O by 3.5, 2.7 and 1.6 AR, and on YCB-V by 3.0, 3.8 and 1.8. ICP leads on TUD-L by 0.4, 3.3 and 0.2 respectively. The preferred refiner is therefore dataset-dependent. The table's average gains are computed directly from each trial's three displayed dataset scores relative to trial 0; the per-dataset results remain the basis for these comparisons.

\subsection{Auxiliary Evaluations}
\label{sec:auxiliary evaluations}

\sectitle{Label noise and test-label sensitivity.}
Real pose annotations can be imperfect, so reducing the synthetic-to-real domain gap need not be the only reason that Z2H improves a student. Depth refinement and verification may also produce pseudo-poses that reduce harmful annotation noise for some instances. The matched-image results in Supplement Table S2 motivate this hypothesis: silver-label supervision exceeds GT supervision for all three students on YCB-V, whereas GT supervision remains stronger on TUD-L. This pattern is consistent with a dataset-dependent contribution from label quality, but does not isolate it from other differences in the supervision signal.

\Cref{tab:relabel} evaluates sensitivity to replacing the test labels on five datasets. The labeling pipeline uses a B2TFPose teacher, PBR-trained GDRNPP and ICP, while student training uses FoundationPose, its refiner and ZebraPose. Different models reduce direct reuse of the training pipeline, but do not make the replacement labels independent ground truth: both pipelines rely on geometric refinement and verification.

Block (a) evaluates all original test targets against BOP ground truth. Block (b) retains only target instances accepted by the replacement-label verifier, while still scoring against original ground truth. All three models improve on this subset. This can reflect easier, more visible or otherwise better-conditioned instances; it does not by itself establish errors in the original labels. No evaluation-target image is used for student training.

Block (c) uses replacement labels on the same instances as (b). Z2H's average falls from 82.8 to 79.7 AR, with lower scores on four datasets and a higher score on LM-O. The largest decline is on T-LESS (89.5 to 76.4). Z2H still exceeds both comparison models in every dataset under both label sets. The preservation of Z2H's advantage under both label sets supports investigating label noise as a contributing factor. However, the experiment cannot identify which labels are more accurate or attribute the changes uniquely to annotation noise. These study-specific baselines are separate from the published benchmark and the matched-label training experiment in Supplement Table S2.

\sectitle{Amount of unlabeled real data.}
\input{figs/fig_unlabeled_vs_AR_plot}
\Cref{fig:analysis} varies the fraction of unlabeled real data used to train ZebraPose in a separate single-seed study (Supplement Table S4). At 5\%, TUD-L improves from 75.2 to 82.8 AR and LM-O from 73.5 to 77.7, gains of 7.6 and 4.2 points. At 25\%, the scores reach 84.6 and 78.6, within 0.2 of the full-data results (84.8 and 78.8). Performance is not strictly monotonic: both scores decline slightly at 50\%. These measurements show that much of the observed benefit is available with a small fraction of the data, without establishing that more data always improves accuracy. Their baselines and seed differ from the main three-seed benchmark.

\input{sec/similarity_table}
\sectitle{When the teacher matters.}
Supplement Table S5 varies the synthetic training budget of the student on TUD-L. At 1\% of that budget, teacher-only supervision reaches 73.4 AR versus 64.7 for self-training, and the ensemble reaches 74.6. The teacher-only advantage is therefore 8.7 points, and the ensemble advantage is 9.9. At 5\%, the ensemble and self-training both reach 83.6, while at the full budget their difference is 0.4. This single-seed evidence suggests that the additional teacher is most useful when the synthetic-trained student is weak.

\sectitle{A second student and verifier quality.}
Supplement Table S6 repeats component ablations with GDRNPP in a separate seed-0 study. Refinement improves all nine dataset/configuration combinations relative to raw pseudo-labels. Verification then improves eight of nine, with a 0.1-point decrease for student-only YCB-V. The verifier's precision is 0.93, 0.98 and 0.96 on LM-O, TUD-L and YCB-V, with recalls of 0.99, 1.00 and 0.96 (Supplement Table S7). These results measure candidate filtering; they do not compare the intrinsic difficulty of verification and pose estimation.

\sectitle{Silver-label quality and coverage.}
\Cref{tab:silver} compares kept silver labels with the existing real-image annotations. On LM-O and TUD-L, respectively, 94.2\% and 98.2\% of kept labels lie within $0.1d$, and 67.1\% and 98.0\% lie within $0.05d$. The reported rotation errors are 3.7 and 2.3 degrees, and translation errors are 6.5 and 3.9\,mm. These measurements establish agreement with the reference annotations, not superiority to them. On LM-O, 27.1\% of kept labels lie between the two thresholds and 5.8\% remain outside $0.1d$; the table does not determine the failure mechanism of those remaining labels.

Coverage also matters. Of all candidate instances, 77\% are kept on LM-O and 73\% on TUD-L; the verifier rejects 19\% and 14\%, while 4\% and 13\% receive no candidate. On LM-O, the mean visible fraction falls from 0.84 for kept instances to 0.59 for rejected instances and 0.31 for those without a candidate. Exclusion is therefore associated with lower visibility. On TUD-L the corresponding fractions are 0.98, 0.97 and 0.97, so visibility alone does not explain exclusion. The results do not isolate an alternative cause such as appearance shift.

\sectitle{Ensembling schemes.}
\cref{tab:ensemble} compares three ways of aggregating the candidates of the Z2H ensemble.
Selecting the candidate with the minimum Chamfer distance to the depth and keeping the union of all verified candidates perform on par (81.8 AR on average), with the former performing better on TUD-L (84.8 vs.\ 84.5 AR), and the latter on LM-O (79.0 vs.\ 78.7 AR).
We have used the minimum Chamfer distance 
as the aggregation method in all experiments,
as it has the added advantage of yielding a smaller
training dataset, containing at most one selected candidate per object and image instead of up to two from a two-teacher union.
Averaging candidate poses performs clearly worse
on both datasets, and
3.8 AR lower on average (78.0 versus 81.8). 

\input{sec/supp_ensemble}

\sectitle{Inference cost.}
Z2H leaves the deployed student's computation unchanged: refinement, verification and ensemble selection are performed only during pseudo-label generation. The supplied timing measurements report more than two minutes per image for the B2TFPose teacher and 0.08\,s per object for GDRNPP. These use different units and do not establish a directly comparable speedup factor. The architectural conclusion is narrower: adapting a student with Z2H adds no teacher or depth-processing stage at inference.

%% file: sec/table_1.tex
\begin{table*}[t]
\centering
\caption{\textbf{Comparison with supervised RGB methods on the seven core BOP datasets:} (AR\,$\uparrow$, BOP19 test targets, official toolkit). All methods are trained on synthetic labeled images $\{X^S,Y^S\}$; the \textit{Labels} columns mark which real images and corresponding labels are also used in training: \textbf{$Y^R$} = real images with ground-truth pose annotations; \textbf{$\widehat{Y}^R$} = real images with pseudo-poses generated by \method. The \textit{SM} column indicates whether a single model is trained for all objects in the dataset,
or whether a distinct model is trained for each object.
Real image depth is used only for pseudo-pose refinement, verification and selection during pseudo-pose generation; student training and inference are RGB-only. \method uses ensembling (\cref{sec:method}), mean of three seeds. ITODD and HB carry the synthetic-only leaderboard result of each student.
$D$: no test-scene images; $T$: adaptation on unlabeled non-target frames from test scenes for LM-O, T-LESS and IC-BIN. The latter setting has additional scene access and is reported separately. IC-BIN in $D$ retains the synthetic-only result. The $D$ LM-O score is the mean of 75.2, 74.5 and 74.8 (sample std 0.35 AR). Parenthesized gains use the listed real-GT baseline for SurfEmb and GDRNPP, and the synthetic-only EffNet-B4 baseline for ZebraPose.  
}
\label{tab:main}
\setlength{\tabcolsep}{3.6pt}
\resizebox{\textwidth}{!}{
\begin{tabular}{l cc c ccccccc c}
\toprule
& \multicolumn{2}{c}{Labels} & & \multicolumn{7}{c}{Dataset} & \\
\cmidrule(lr){2-3}\cmidrule(lr){5-11}
Method & $Y^R$ & $\widehat{Y}^R$ & SM & LM-O & T-LESS & TUD-L & IC-BIN & ITODD & HB & YCB-V & Avg \\
\midrule
CosyPose~\cite{labbe2020cosypose} & \xmrk & \xmrk & \xmrk & 63.3 & 64.0 & 68.5 & 58.3 & 21.6 & 65.6 & 57.4 & 57.0 \\
CDPNv2~\cite{li2019cdpn} & \xmrk & \xmrk & \xmrk & 62.4 & 40.7 & 58.8 & 47.3 & 10.2 & 72.2 & 39.0 & 47.2 \\
DPODv2~\cite{shugurov2021dpodv2} & \xmrk & \xmrk & \xmrk & 58.4 & 63.6 & -- & -- & -- & 72.5 & -- & -- \\
EPOS~\cite{hodan2020epos} & \xmrk & \xmrk & \chk & 54.7 & 46.7 & 55.8 & 36.3 & 18.6 & 58.0 & 49.9 & 45.7 \\
SurfEmb~\cite{haugaard2022surfemb} & \xmrk & \xmrk & \chk & 65.6 & 74.1 & 71.5 & 58.5 & 38.7 & 79.3 & 65.3 & 64.7 \\
ZebraPose~\cite{su2022zebrapose} & \xmrk & \xmrk & \xmrk & 72.1 & 72.3 & 71.7 & 54.5 & 41.0 & 88.2 & 69.1 & 67.0 \\
ZebraPose (EffNet-B4)~\cite{su2022zebrapose} & \xmrk & \xmrk & \xmrk & 72.9 & 81.1 & 75.6 & 59.2 & \underline{50.4} & \underline{92.1} & 72.9 & 72.0 \\
PFA~\cite{hu2022pfa} & \xmrk & \xmrk & \chk & 74.5 & 71.9 & 73.2 & 60.0 & 35.3 & 84.1 & 64.8 & 66.3 \\
GDRNPP~\cite{liu2025gdrnpp} & \xmrk & \xmrk & \xmrk & 71.3 & 79.6 & 75.2 & 62.3 & 44.8 & 86.9 & 71.3 & 70.2 \\
\hline
Pix2Pose~\cite{park2019pix2pose} & \chk & \xmrk & \xmrk & 36.3 & 34.4 & 42.0 & 22.6 & 13.4 & 44.6 & 45.7 & 34.2 \\
CDPNv2~\cite{li2019cdpn} & \chk & \xmrk & \xmrk & 62.4 & 47.8 & 77.2 & 47.3 & 10.2 & 72.2 & 53.2 & 52.9 \\
CRT-6D~\cite{castro2023crt6d} & \chk & \xmrk & \chk & 66.0 & 64.4 & 78.9 & 53.7 & 20.8 & 60.3 & 75.2 & 59.9 \\
CosyPose~\cite{labbe2020cosypose} & \chk & \xmrk & \chk & 63.3 & 72.8 & 82.3 & 58.3 & 21.6 & 65.6 & 82.1 & 63.7 \\
SurfEmb~\cite{haugaard2022surfemb} & \chk & \xmrk & \chk & 65.6 & 77.0 & 80.5 & 58.5 & 38.7 & 79.3 & 71.8 & 67.3 \\
GDRNPP (SM)~\cite{liu2025gdrnpp} & \chk & \xmrk & \chk & 68.6 & 77.6 & 82.7 & 61.7 & 26.0 & 80.9 & 76.8 & 67.8 \\
PFA~\cite{hu2022pfa} & \chk & \xmrk & \chk & 74.5 & 77.8 & 83.9 & 60.0 & 35.3 & 84.1 & 80.6 & 70.9 \\
ZebraPose~\cite{su2022zebrapose} & \chk & \xmrk & \xmrk & 72.1 & 80.6 & 85.0 & 54.5 & 41.0 & 88.2 & 83.0 & 72.0 \\
GDRNPP~\cite{liu2025gdrnpp} & \chk & \xmrk & \xmrk & 71.3 & 78.6 & 83.1 & 62.3 & 44.8 & 86.9 & 82.5 & 72.8 \\
ZebraPoseSAT~\cite{su2022zebrapose} & \chk & \xmrk & \xmrk & 72.9 & \underline{82.1} & 85.0 & 59.2 & \underline{50.4} & \textbf{92.2} & 82.8 & 74.9 \\
HccePose(BF)~\cite{wang2025hccepose} & \chk & \xmrk & \xmrk & 75.5 & \textbf{85.6} & \textbf{86.9} & 63.5 & \textbf{54.2} & 91.9 & \underline{83.9} & \underline{77.3} \\
\hline
\method{}:FP/SurfEmb ($T$)
& \xmrk & \chk & \chk & 74.7 & 77.0 & 82.4 & 64.4 & 38.7 & 79.3 & 77.6 & 70.6 (+3.3) \\
\method{}:FP/GDRNPP ($T$)
& \xmrk & \chk & \xmrk & \underline{77.2} & 81.4 & 82.0 & \underline{65.6} & 44.8 & 86.9 & 82.5 & 74.3 (+1.5)\\
\method{}:FP/ZebraPose ($D$)
& \xmrk & \chk & \xmrk & 74.8 & 80.5 & \underline{86.8} & 59.2 & \underline{50.4} & \underline{92.1} & \textbf{84.1} & 75.4 (+3.4) \\
\method{}:FP/ZebraPose ($T$)
& \xmrk & \chk & \xmrk & \textbf{78.8} & \textbf{85.6} & \underline{86.8} & \textbf{65.8} & \underline{50.4} & \underline{92.1} & \textbf{84.1} & \textbf{77.7} (+5.7) \\
\bottomrule
\end{tabular}}
\end{table*}

%% file: figs/qualitative_results.tex
\begin{figure*}[t]
\centering
\includegraphics[width=\textwidth]{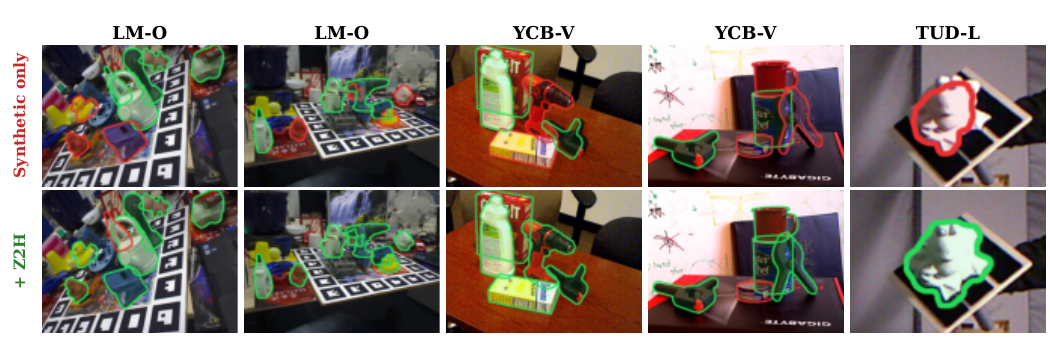}
\vspace{-5mm}
\caption{\textbf{Qualitative results.} Predicted poses 
on BOP test images
of the synthetic image
trained model (top) and after adaptation with \method (bottom); green: correct within $0.1d$, red: wrong.}
\label{fig:qual}
\end{figure*}

%% file: sec/table_new_comparisons.tex
\begin{table}[t]
\centering\small
\caption{\textbf{Domain adaptation on LM-O.} ADD(-S) (\%), Self6D++ protocol. Z2H uses B2TFPose and PBR-trained GDRNPP teachers, the FoundationPose refiner and a GDRNPP student. Baselines are reported comparison values.}
\label{tab:labelfree}
\begin{tabular}{lr}\toprule
Method & ADD(-S) \\\midrule
SMOC-Net~\cite{tan2023smocnet} & 63.3 \\
Self6D++~\cite{wang2021self6dpp} & 64.7 \\
PseudoFlow~\cite{hai2023pseudoflow} & 65.4 \\
TexPose~\cite{chen2023texpose} & 66.7 \\
RKHSPose~\cite{wu2024rkhspose} & 68.6 \\
ONDA-Pose~\cite{tan2025ondapose} & 69.3 \\
Z2H & \textbf{70.4} \\
\bottomrule\end{tabular}
\end{table}

%% file: sec/ablation_table.tex
\begin{table*}[t]
\centering
\caption{Ablations: Impact of refinement and verification
modules on \method AR\,$\uparrow$, with ZebraPose (ZP) student,
using either ICP or FoundationPose as refiner.
Trials 1--9 use B2TFPose (B2TF) wherever a zero-shot teacher is included.
Trial 10 replaces B2TFPose with FoundationPose (FP) and corresponds
to the full method's test-scene adaptation setting in
\cref{tab:main}.
$\overline{\Delta}$ reports the mean of the three displayed scores minus the mean of trial 0 (73.9), rounded to one decimal.
--: not run.}
\label{tab:ablation}
\setlength{\tabcolsep}{4pt}
\begin{tabular}{c c cc cc cccc cccc}
\toprule
 & 
 &
 \multicolumn{2}{c}{Ensemble}
 & \multicolumn{2}{c}{}
 & \multicolumn{4}{c}{ICP refiner}
 & \multicolumn{4}{c}{FoundationPose refiner} \\
\cmidrule(lr){3-4}
\cmidrule(lr){7-10}
\cmidrule(lr){11-14}
Trial \# & 
Teacher 
&
Zero-shot & PBR
 & Refine & Verify
 & LM-O & YCB-V & TUD-L & $\overline{\Delta}$
 & LM-O & YCB-V & TUD-L & $\overline{\Delta}$ \\
\midrule
0 & 
--
&
\xmrk & \xmrk & \xmrk & \xmrk
  & 72.9 & 73.0 & 75.8 & 0 & 72.9 & 73.0 & 75.8 & 0 \\
1 &
B2TF
& \chk & \xmrk & \xmrk & \xmrk
  & 71.3 & 70.8 & 77.3 & -0.8 & 71.3 & 70.8 & 77.3 & -0.8 \\
2 &
B2TF
& \chk & \xmrk & \chk & \xmrk
  & 72.3 & 80.1 & 83.3 & +4.7 & 77.7 & 84.4 & 83.7 & +8.0 \\
3 &
B2TF
& \chk & \xmrk & \chk & \chk
  & 75.0 & 80.9 & 84.3 & +6.2 & 78.5 & 83.9 & 83.9 & +8.2 \\
4 &
ZP
& \xmrk & \chk & \xmrk & \xmrk
  & 71.1 & 70.9 & 78.0 & -0.6 & 71.1 & 70.9 & 78.0 & -0.6 \\
5 &
ZP
& \xmrk & \chk & \chk & \xmrk
  & 72.2 & 77.9 & 83.5 & +4.0 & 74.0 & 82.0 & 78.4 & +4.2 \\
6 &
ZP
& \xmrk & \chk & \chk & \chk
  & 73.9 & 79.4 & 83.9 & +5.2 & 76.6 & 83.2 & 80.6 & +6.2 \\
7 &
B2TF, ZP
& \chk & \chk & \xmrk & \xmrk
  & 73.9 & 78.1 & 80.8 & +3.7 & 73.9 & 78.1 & 80.8 & +3.7 \\
8 &
B2TF, ZP
& \chk & \chk & \chk & \xmrk
  & 74.3 & 80.7 & 84.3 & +5.9 & 75.0 & 81.9 & 75.7 & +3.6 \\
9 &
B2TF, ZP
& \chk & \chk & \chk & \chk
  & 76.2 & 81.8 & 84.5 & +6.9 & 77.8 & 83.6 & 84.3 & +8.0 \\
10 &
FP, ZP
& \chk & \chk & \chk & \chk
   & -- & -- & -- & --
   & 78.8 & 84.1 & 86.8 & +9.3 \\
\bottomrule
\end{tabular}
\label{table:ablations}
\end{table*}

%% file: sec/supp_relabel.tex
\begin{table*}[t]
\centering
\caption{\textbf{Test data re-annotation} (AR\,$\uparrow$, ZebraPose student, BOP19 test targets, official toolkit). The test images are re-annotated by a different Z2H pipeline (B2TFPose teacher, PBR-trained GDRNPP, ICP refiner), different from the pipeline that trains \method (FoundationPose teacher and refiner, ZebraPose). All models are trained on synthetic PBR images with their ground-truth poses $\{X^S,Y^S\}$. The \textit{Labels} columns mark which labels on \emph{real} training images are used in addition: \textbf{$Y^R$} = real training images with their ground-truth pose annotations (manual dataset labels); \textbf{$\widehat{Y}^R$} = real training images with pseudo-poses generated automatically by \method, without using any ground truth. In this study, the fully supervised reference for LM-O and IC-BIN is the PBR model. (a) original BOP ground truth, all targets; (b) original ground truth, restricted to the instances kept by verification; (c) replacement test labels on the same instances as (b). \textit{Kept}: share of test instances that pass verification. \textbf{Bold}: best per column within each block.}
\label{tab:relabel}
\setlength{\tabcolsep}{5pt}
\begin{tabular}{l cc ccccc c}
\toprule
& \multicolumn{2}{c}{Labels} & \multicolumn{5}{c}{Dataset} & \\
\cmidrule(lr){2-3}\cmidrule(lr){4-8}
Model & $Y^R$ & $\widehat{Y}^R$ & LM-O & T-LESS & TUD-L & IC-BIN & YCB-V & Avg \\
\midrule
\multicolumn{9}{l}{\textit{(a) GT poses, all test targets}} \\
\quad PBR only         & \xmrk & \xmrk & 72.9 & 81.0 & 75.8 & 61.2 & 73.0 & 72.8 \\
\quad Fully supervised & \chk  & \xmrk & 72.9 & 81.1 & 83.7 & 61.2 & 82.8 & 76.3 \\
\quad \method          & \xmrk & \chk  & \textbf{78.8} & \textbf{85.6} & \textbf{86.8} & \textbf{65.8} & \textbf{84.1} & \textbf{80.2} \\
\midrule
\multicolumn{9}{l}{\textit{(b) GT poses, kept test target subset}} \\
\quad PBR only         & \xmrk & \xmrk & 75.7 & 85.5 & 76.7 & 66.3 & 74.4 & 75.7 \\
\quad Fully supervised & \chk  & \xmrk & 75.7 & 85.9 & 84.4 & 66.3 & 84.9 & 79.4 \\
\quad \method          & \xmrk & \chk  & \textbf{81.8} & \textbf{89.5} & \textbf{87.4} & \textbf{69.1} & \textbf{86.0} & \textbf{82.8} \\
\midrule
\multicolumn{9}{l}{\textit{(c) Pseudo-poses, kept test target subset}} \\
\quad PBR only         & \xmrk & \xmrk & 78.1 & 71.2 & 75.7 & 66.2 & 71.9 & 72.6 \\
\quad Fully supervised & \chk  & \xmrk & 78.1 & 71.2 & 81.6 & 66.2 & 81.7 & 75.8 \\
\quad \method          & \xmrk & \chk  & \textbf{83.8} & \textbf{76.4} & \textbf{87.1} & \textbf{68.0} & \textbf{83.1} & \textbf{79.7} \\
\midrule
Kept (\%) & & & 83 & 77 & 93 & 50 & 88 & -- \\
\bottomrule
\end{tabular}
\end{table*}

%% file: figs/fig_unlabeled_vs_AR_plot.tex
\begin{figure}[t]
\centering
\includegraphics[width=\linewidth]{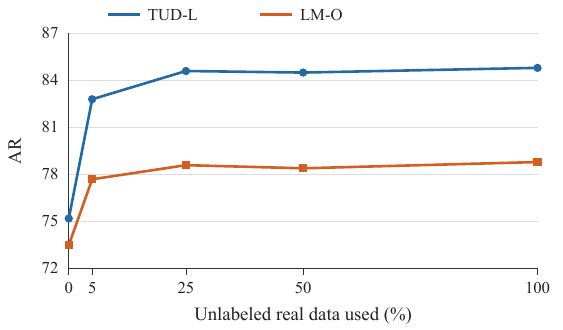}
\vspace{-1mm}
\caption{Z2H:FP/ZebraPose AR as a function of the fraction of unlabeled real data used for TUD-L and LM-O datasets. 0\% = synthetic-only baselines. Full values provided in Supp. Tab. S4.}
\label{fig:analysis}
\end{figure}

%% file: sec/similarity_table.tex
\begin{table*}[t]
\centering
\caption{\textbf{Silver pseudo-pose quality and coverage} on the unlabeled real training images (FoundationPose teacher and refiner, verified). Ground truth is used only for this analysis, never for training or label selection. \textit{Candidates}: object instances in the unlabeled images; each is either kept as a silver label, rejected by the verifier, or receives no candidate pose. Accuracy columns refer to the kept silver labels; $d$ = object diameter; within $0.1d$ / $0.05d$: ADD(-S) error below 10\% / 5\% of $d$. \textit{Visibility}: mean visible fraction of the object for kept / rejected / no-candidate instances.}
\label{tab:silver}
\setlength{\tabcolsep}{4pt}
\resizebox{\textwidth}{!}{
\begin{tabular}{l c ccc ccccc c}
\toprule
& & \multicolumn{3}{c}{Coverage} & \multicolumn{5}{c}{Accuracy of kept labels} & \\
\cmidrule(lr){3-5}\cmidrule(lr){6-10}
Dataset & Candidates & Kept & Rejected & No cand. & within $0.1d$ & within $0.05d$ & ADD(-S)$/d$ & Rot.\,($^\circ$) & Trans.\,(mm) & Visibility \\
\midrule
LM-O  & 7{,}670  & 5{,}916 (77\%) & 1{,}467 (19\%) & 287 (4\%)  & 94.2\% & 67.1\% & 0.043 & 3.7 & 6.5 & 0.84 / 0.59 / 0.31 \\
TUD-L & 2{,}135  & 1{,}568 (73\%) & 295 (14\%)     & 272 (13\%) & 98.2\% & 98.0\% & 0.017 & 2.3 & 3.9 & 0.98 / 0.97 / 0.97 \\
\bottomrule
\end{tabular}}
\vspace{-2mm}
\end{table*}

%% file: sec/supp_ensemble.tex
\begin{table}[t]
\centering
\caption{\textbf{Ensembling schemes} (AR\,$\uparrow$, ZebraPose student, FoundationPose teacher and refiner, verified, single seed). \textit{Min Chamfer}: per object and image, keep candidate with minimum Chamfer distance to observed depth (ours); \textit{Union}: keep every verified candidate; \textit{Averaging}: average of candidate poses. 
}
\label{tab:ensemble}
\setlength{\tabcolsep}{6pt}
\begin{tabular}{l cc c}
\toprule
Ensembling & LM-O & TUD-L & Avg \\
\midrule
Min Chamfer (ours) & 78.7 & \textbf{84.8} & \textbf{81.8} \\
Union              & \textbf{79.0} & 84.5 & \textbf{81.8} \\
Averaging          & 72.5 & 83.4 & 78.0 \\
\bottomrule
\end{tabular}
\vspace{-2mm}
\end{table}

%% file: sec/5_conclusion.tex
\section{Conclusion}
Z2H combines zero-shot and simulation-trained teachers with geometric refinement, verification and hard label selection to adapt RGB pose estimators without real pose annotations. On the seven core BOP datasets, ZebraPose reaches 75.4 AR without test-scene imagery and 77.7 with unlabeled non-target test-scene frames, compared with its 72.0 synthetic-only baseline. SurfEmb and GDRNPP also improve, and the LM-O domain-adaptation comparison reaches 70.4 ADD(-S). The ablations support refinement and verification as important components, while showing that the best teacher and refiner depend on the dataset. The deployed student's architecture and RGB-only inference remain unchanged. The matched-label and re-annotation studies also motivate a label-quality hypothesis: refined, verified pseudo-poses may reduce harmful annotation noise for some datasets, alongside reducing the domain gap. Establishing the size of that contribution requires independent pose-label validation.

\sectitle{Limitations.}
Z2H requires unlabeled real RGB-D data, offline teacher computation and student training. Its silver labels are imperfect and omit some instances, especially less visible ones on LM-O. Benchmark comparisons do not fully control scene access and training data. Matched-image experiments show that silver supervision can exceed GT supervision on one dataset and remain below it on another. Test re-annotation is a sensitivity analysis, not an independent validation of annotation accuracy. Single-seed auxiliary studies do not establish variability across training runs.

\sectitle{Societal impact.}
Reducing manual annotation may speed up deployment and lower development costs. Inaccurate poses can affect downstream robotic decisions, so deployment requires application-specific evaluation.